\documentclass{article}

\usepackage{xcolor}
\usepackage{subfig}
\usepackage{adjustbox}
\usepackage{amsmath, amssymb}
\usepackage{subcaption}
\usepackage{arxiv}

\usepackage[utf8]{inputenc} 
\usepackage[T1]{fontenc}    
\usepackage{hyperref}       
\usepackage{url}            
\usepackage{booktabs}       
\usepackage{amsfonts}       
\usepackage{nicefrac}       
\usepackage{microtype}      
\usepackage{lipsum}
\usepackage{graphicx}
\graphicspath{ {./images/} }

\newcommand{\subfigwidth}{0.23\textwidth} 

\title{MIHRaGe: A Mixed-Reality Interface for Human-Robot Interaction via Gaze-Oriented Control}

\author{
  Rafael R. Baptista\\
  Engineering School of São Carlos\\
  University of São Paulo \\
  São Carlos, Brazil\\
  \And
  Nina R. Gerszberg\\
  Massachussets Institute of Technology\\
  Boston, USA\\
  \And
  Ricardo V. Godoy\\
  Engineering School of São Carlos\\
  University of São Paulo \\
  São Carlos, Brazil\\
  \And
  Gustavo J. G. Lahr \\
  Hospital Israelita Albert Einstein \\
  São Paulo, Brazil\\
}

\begin{document}
\maketitle
\begin{abstract}
Individuals with upper limb mobility impairments often require assistive technologies to perform activities of daily living. While gaze-tracking has emerged as a promising method for robotic assistance, existing solutions lack sufficient feedback mechanisms, leading to uncertainty in user intent recognition and reduced adaptability. This paper presents the MIHRAGe interface, an integrated system that combines gaze-tracking, robotic assistance, and a mixed-reality to create an immersive environment for controlling the robot using only eye movements. The system was evaluated through an experimental protocol involving four participants, assessing gaze accuracy, robotic positioning precision, and the overall success of a pick and place task. Results showed an average gaze fixation error of 1.46 cm, with individual variations ranging from 1.28 cm to 2.14 cm. The robotic arm demonstrated an average positioning error of $\leq$1.53 cm, with discrepancies attributed to interface resolution and calibration constraints. In a pick and place task, the system achieved a success rate of 80\%, highlighting its potential for improving accessibility in human-robot interaction with visual feedback to the user.
\end{abstract}

\section{Introduction}
Subjects that suffer from conditions that affect upper limb mobility, such as spinal cord injuries, neurodegenerative diseases, and strokes, often rely on external assistance due to the significant level of impairment to perform activities of daily living (ADLs)~\cite{cattaneo2017participationSclerosis, anwer2022rehabilitationStroke}. Assistive robotic systems have been developed to address this challenge, including exoskeletons, prosthetics, orthotics, and supernumerary robotic limbs, which can replace or augment lost motor functions~\cite{Khalid2023rehabRobotics, Hasan2024UpperLimbRehab}. These approaches leverage biosignals, including voice, head, or mouth movements, but most assistive robotic methods have inherent limitations, such as limb fatigue~\cite{jing2024mouthFatigue} and high susceptibility to background noise~\cite{saood2024designing,wang2024multimodal}. Eye tracking technologies have emerged as a promising solution, enabling real-time intention decoding and robot-actuated assistance through gaze-based control without dealing with limb fatigue and less background noise sensitivity~\cite{raju2024signal, sunny2021eye, fischer2024scoping}.

\begin{figure}
    \centering
    \includegraphics[width=0.98\linewidth]{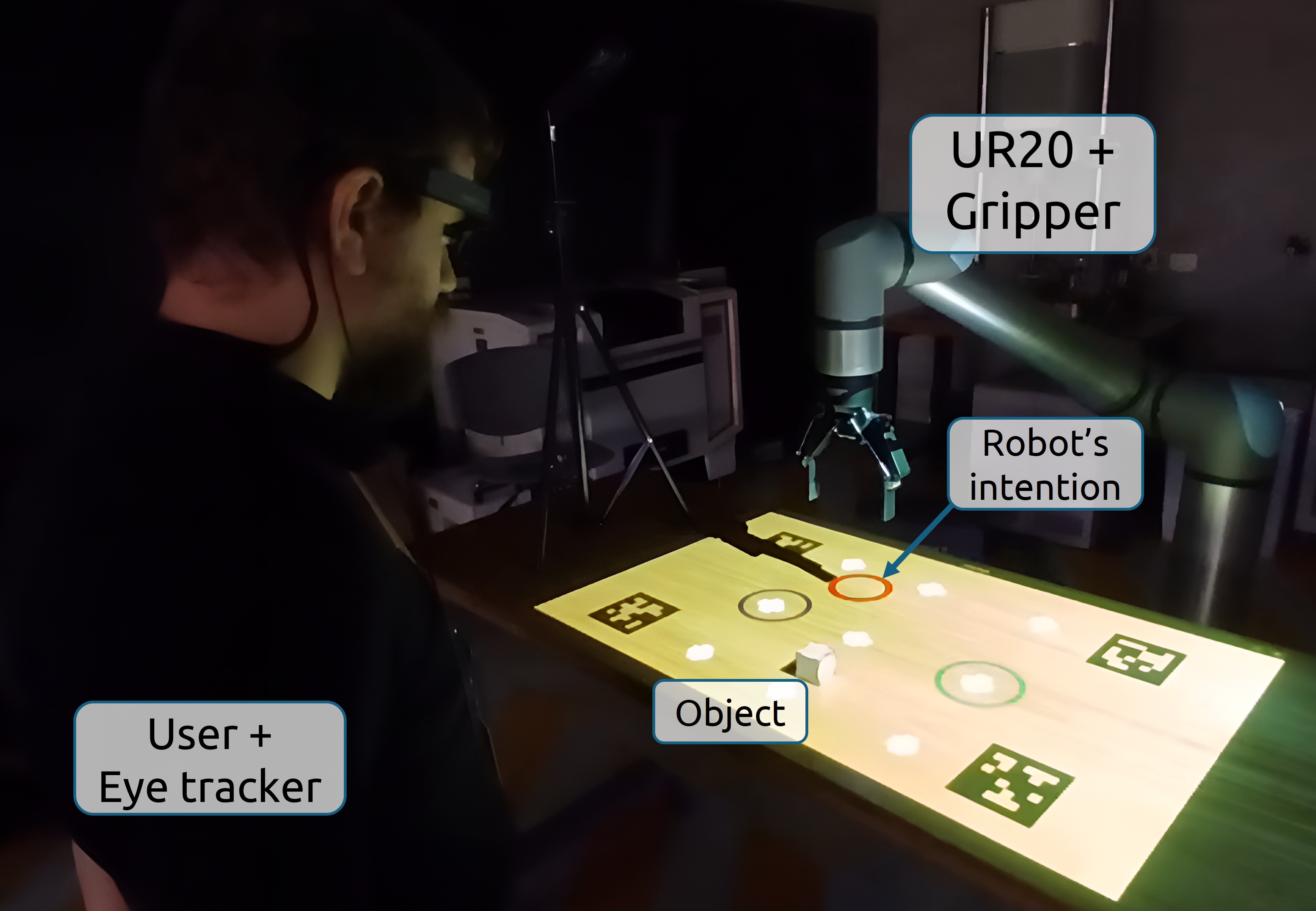}
    \caption{Overview of the MIHRAGe interface: a user equipped with an eye tracker controls a robotic arm through a mixed-reality interface, which shows user control and robot intent.}
    \label{fig:intro}
\end{figure}

Several studies have integrated human-robot interaction with eye-tracking technologies to infer user intent and facilitate hands-free object manipulation. These systems leverage gaze-based selection mechanisms to generate movement commands, allowing users to interact with robotic arms and other assistive devices solely through eye movements~\cite{belardinelli2024gaze}. Research has demonstrated the effectiveness of gaze-driven control across a wide range of applications, including assistive pick-and-place tasks, autonomous object retrieval, and remote robotic teleoperation, providing an alternative to conventional input methods that rely on residual motor function~\cite{cambuzat2018immersive, fuchs2021gaze, davalos20243d}.

Despite these advancements, a significant gap remains in user experience limitations, particularly regarding the clarity of feedback in gaze-controlled robotic systems. One of the primary challenges is the lack of explicit feedback from the robot regarding its intended actions, which can lead to uncertainty in human-robot interaction~\cite{kim2020errors, axelsson2022modeling}. Users often struggle to determine whether their gaze input has been correctly registered and interpreted by the system, which can reduce confidence and trust in its functionality~\cite{saran2018human, prajod2023gaze}. Studies have shown that ambiguous system responses increase cognitive load, forcing users to repeat trial-and-error interactions to confirm the robot’s comprehension of their intent~\cite{sonawani2022HeniMixedReality}. Recent research suggests that multimodal feedback mechanisms, such as augmented reality overlays or vibrotactile feedback, could enhance user awareness of robot state and intention, ultimately improving interaction efficiency and reducing user frustration~\cite{walker2023virtual}. However, the implementation of such features remains underexplored in gaze-based robotic control systems, highlighting the need for further research into optimizing feedback strategies to enhance user experience and system reliability.

This paper presents the MIRAHGe interface, a novel approach that integrates gaze-tracking with a robotic arm and a mixed-reality interface. Unlike existing solutions, MIHRAGe provides real-time visual feedback on the gaze selection process and the robot's intended actions, addressing the critical gap in user feedback mechanisms. By leveraging gaze-based control within a projected interface, the system allows users to control the robot's position and execute pick and place with objects without requiring physical interaction. By dynamically overlaying relevant information onto the interface, such as gaze fixation points and robot movement intent, mixed reality helps users better anticipate and visualize system response.

\subsection{Experimental setup}

The MIHRAGe interface uses a Tobii Pro Glasses 2 as the eye tracking device, which provides gaze coordinates in pixels at its frame, and also has a front-facing camera that was utilized to recognize the visual interface's markers (AprilTags), assuming the user is always looking at the interface. The robotic arm integrated into the system was a Universal Robots UR20, a six-degree-of-freedom robot equipped with a Robotiq parallel gripper model 2F-140 for object manipulation. The interface, programmed using OpenCV-Python, is shown to the user using an Epson projector Powerlite E20. An overview of the experimental setup is shown in Fig.~\ref{fig:intro}. All the equipment is connected to a ROS2 network.


The distance between the center of the markers is 70 cm $\times$ 38.5 cm, and the area for interaction (i.e., the workspace) has 57 cm $\times$ 25.6 cm. The interface was positioned 60~cm from the user to maintain comfortable visibility and effective gaze tracking. The system was powered by an Avell A52 laptop with processor i7-12650H, 32 GB of RAM, GPU GeForce RTX 3050, and Ubuntu 22.04.

\subsection{Mixed Reality Interface}
\begin{figure}
    \centering
    \includegraphics[width=0.98\linewidth]{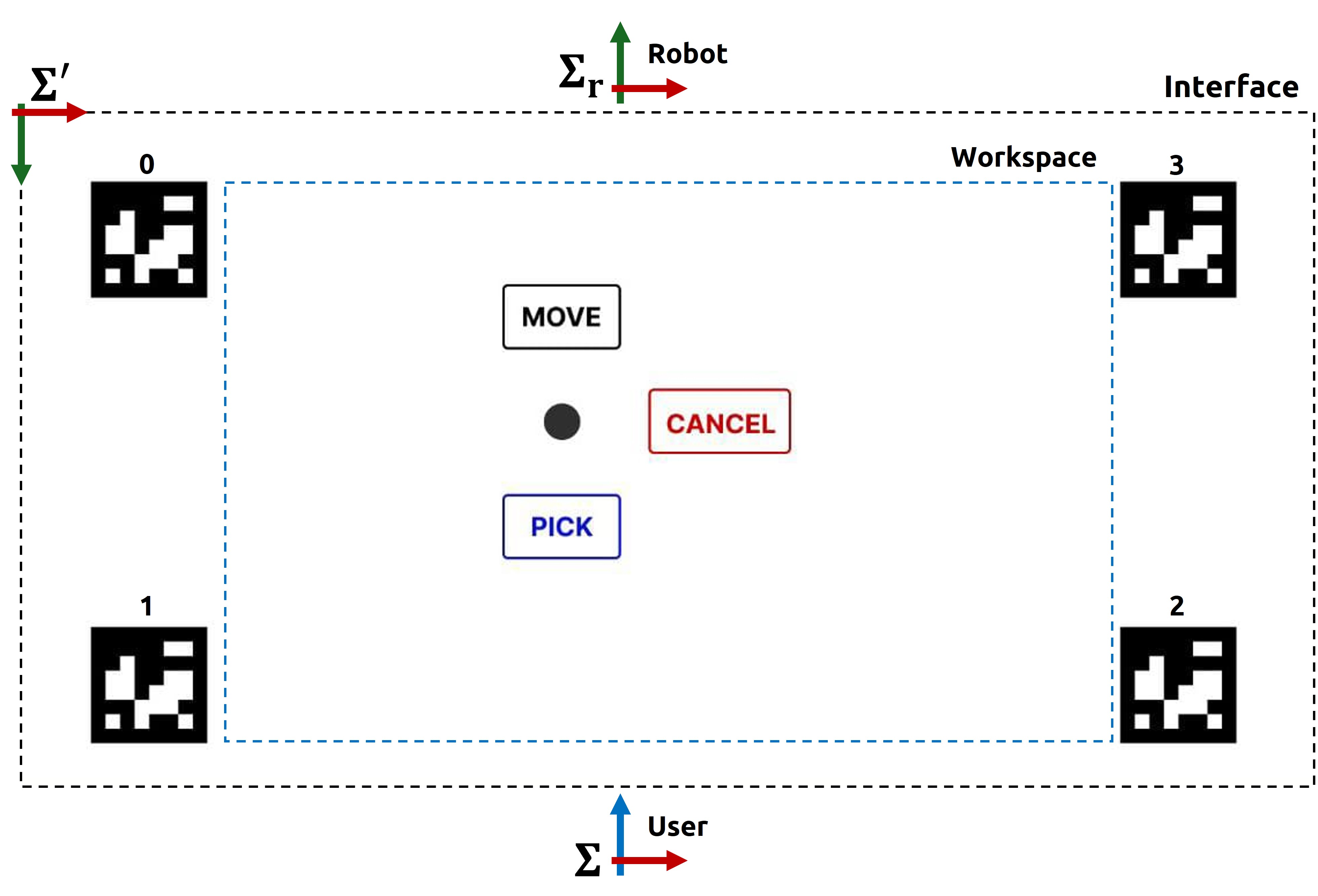}
    \caption{An overview of the interface used to create the mixed-reality environment and the main frames. The interface frame is represented by $\Sigma'$, the robot frame by $\Sigma_r$, and the user frame/eye tracker's frame by $\Sigma$. The workspace for interaction is placed between the four markers (numbered from 0 to 3), along with the menu displayed with the option to choose \textit{PICK}, \textit{MOVE}, and \textit{CANCEL} intention. The dashed lines, the frame orientation axis, and the marker numbering were included for illustrative purposes only.}
    \label{fig:interface}
\end{figure}

The visual guide interface was developed to enable the user to control the movement of the robotic arm solely through gaze. This interface, shown in Fig.~\ref{fig:interface}, facilitates the selection of desired points within a predefined working area. 
It plays a crucial role by providing three main functionalities: user feedback, robot intent visualization, and environment interaction.

The user gazes at the interface while wearing the eye tracker. 
After two seconds of staring at a point, a cursor appears as a gray circle and remains static, marking the selected point. A menu appears around the circle with three options: \textit{MOVE}, which commands the robotic arm to move above the selected point and remain stationary; \textit{PICK}, positioning the robot above the desired point and proceeding to grasp the object using its parallel gripper; and \textit{CANCEL}, used to cancel if the selected point is not of interest. If the \textit{PICK} action is chosen, the next interaction cycle will replace this option with \textit{PLACE}, allowing the user to place the object at another designated location.

\begin{figure}
    \centering
    \includegraphics[width=0.98\linewidth, trim={1.5cm 0 0.98cm 0},clip]{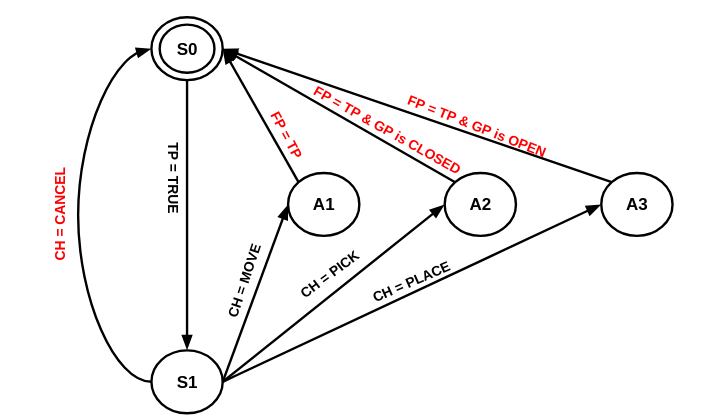}
    \caption{State machine representing the different behaviors adopted by the robot, depending on the choice (CH) value the user selects. TP stands for Target Point, GP is the Gripper, and FP corresponds to Final Point.}
    \label{fig:state_machine}
\end{figure}

The interface workflow and possible user selections are represented through a state machine in Fig. \ref{fig:state_machine}, which outlines the system’s behavior based on the value of the variable \textit{CH}.

\begin{table}
    \centering
    \caption{Description of the states in the finite state machine of Fig. \ref{fig:state_machine}}
    \begin{adjustbox}{width=.65\linewidth}
        \begin{tabular}{lp{5cm}}
            Symbol: State & Description \\ \hline
            S0: Observation      & User looking at the interface without fixing at a point of interest \\
            S1: Menu     & Point of interest is detected; menu appears            \\
            A1: Move   & The robot moves its end-effector to the selected point of interest without moving downwards            \\
            A2: Pick   & The robot moves its end-effector to the selected point of interest, stops right above the object, then moves downwards to pick the object, closes the gripper and moves upwards            \\
            A3: Place  & The robot moves its end-effector to the selected point of interest, stops right above the area, then moves downwards to place the object, opens the gripper and moves upwards            \\ \hline
        \end{tabular}
        \label{tab:state_machine}
    \end{adjustbox}
\end{table}

\subsection{Interaction Between Visual Interface and Eye Tracker}

With the roles of the eye tracker and the visual interface defined, their interaction is crucial for properly functioning the developed system. Since the desired movement point is selected through the combined operation of these two tools, ensuring their seamless integration is essential.

The four markers, along with the exact pixel coordinates of their locations and the gaze data, are processed at the resolution of the eye tracker’s camera. However, an accurate conversion between the camera and interface resolution is required to ensure compatibility with the graphical interface. We use homography matrices since they maintain precision in gaze tracking, regardless of the user's position relative to the observed interface~\cite{corke2011robotics}. These matrices continuously reconstruct the image perspective based on the detected markers.

When the user's forehead camera identifies at least one of the four markers, the system corrects the perspective by aligning the detected marker with its corresponding position on the interface. The interface is continuously reconstructed within the eye tracker’s image frame according to the detected marker using Equation~\eqref{eq:homography_transform}.

\begin{equation}
    \mathbf{p}_1' = \mathbf{H} \mathbf{p}_2
    \label{eq:homography_transform}
\end{equation}

The vector \( p_1' \) contains the locations of the corners of each marker in the generated interface written in frame $\Sigma'$. In contrast, \( p_2 \) is continuously updated with the corner locations of each marker, located in the world frame $\Sigma$. And $\mathbf{H} = [h_{i,j}]$ is the homography matrix with elements $h_{i,j}$ and dimensions $3 \times 3$.

The homography matrix is computed based on its corrected perspective, ensuring that the pixels at the edges of the detected marker match their intended positions in the interface. Finally, the gaze coordinates ($g'$) are determined through this perspective transformation, utilizing the previously computed homography matrix.

\begin{equation}
\mathbf{g}' = \mathbf{H} \mathbf{g}
\end{equation}

Where $\mathbf{g}'=[x'_{\text{gaze}}, y'_{\text{gaze}}, w'_{\text{gaze}}]^T$ is the gaze in $\Sigma'$ frame, and $\mathbf{g}=[x_{\text{gaze}}, y_{\text{gaze}}, 1]^T$ is the gaze in eye tracker's frame.

\subsection{Interaction Between Visual Interface and Robot}

With the gaze mapped to the $\Sigma'$ frame, the user can perform the desired actions outlined in the state machine in fig.~\ref{fig:state_machine} and select coordinates to move the robot. The transform between the desired point in the interface's frame $\mathbf{g}'$ to the same point the robot's frame, $\mathbf{p_t}$, is given by \eqref{eq:desired-point-to-robot-frame}.

\begin{equation}
\mathbf{p_t} = \mathbf{A} \mathbf{g}'
\label{eq:desired-point-to-robot-frame}
\end{equation}

where

\begin{equation}
\mathbf{A}
=
\begin{bmatrix} 
\alpha_x & 0 & -\alpha_x \left( w_i + \frac{w_{m}}{2} \right) \\ 
0 & \alpha_y & -\alpha_y \left( h_i + \frac{h_{m}}{2} \right) \\ 
0 & 0 & 1 
\end{bmatrix}
\end{equation}

\begin{figure}[t]
    \centering
    \includegraphics[width=0.98\linewidth]{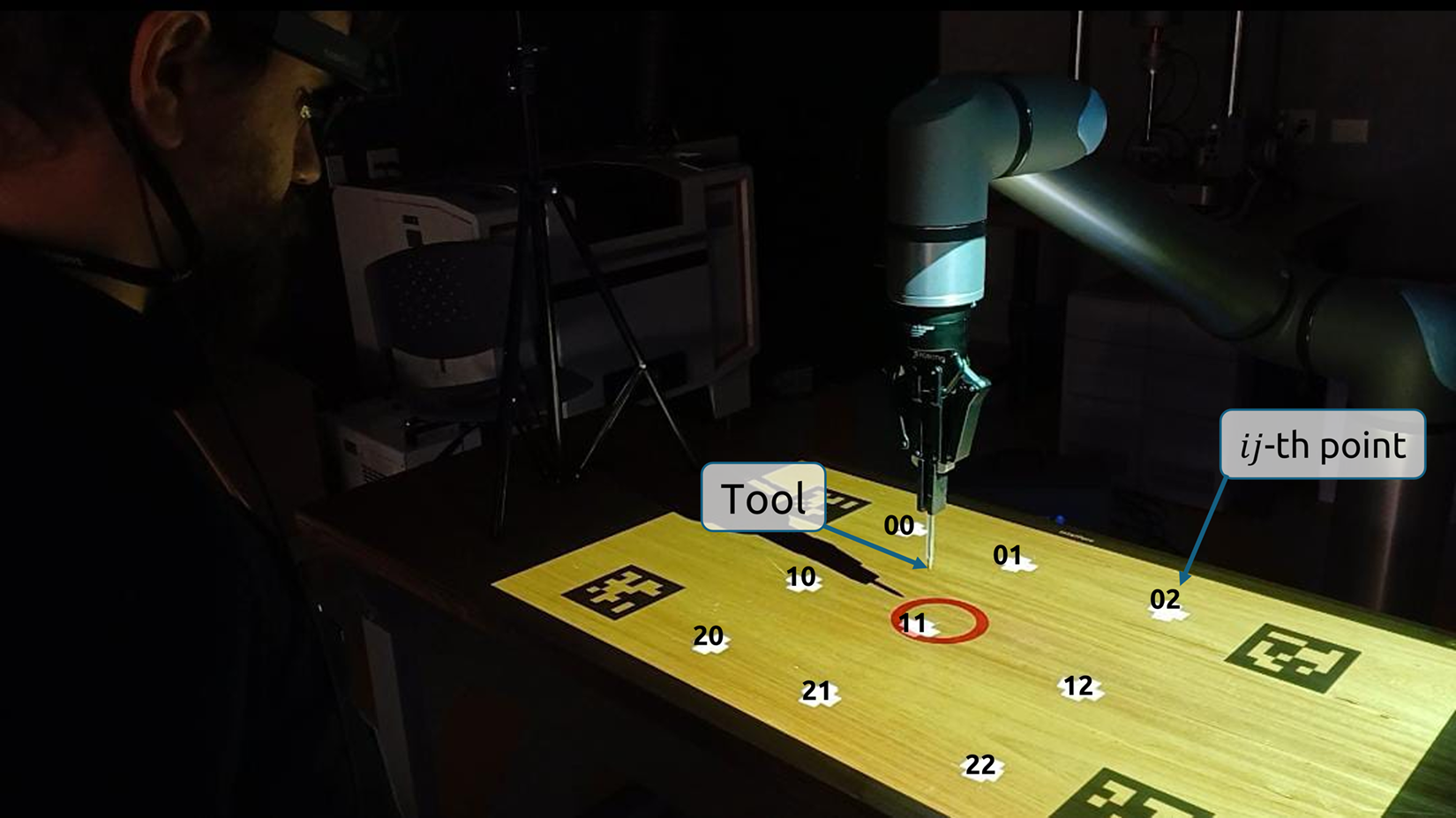}
    \caption{Experiment for gaze accuracy and robot position evaluation. The user fixes their gaze on specific points on the projection surface and selects \textit{MOVE} to pass the target to the robot. The robot then moves to the surface and lowers the tool, enabling the measurement.}
    \label{fig:gaze_evaluation}
\end{figure}

The conversion factor $\alpha$ relates the real-world distance measured from center to center between two markers to the corresponding pixel distance in the generated visual interface. For the x axis, $ \alpha_x = \frac{CC_{x_r}}{CC_{x}} $, where $CC_{x_r}$ is the center-to-center distance between markers in the world frame (chosen to be the same as the robot's), and $CC_{x}$ is the center-to-center distance in pixels in the interface. Additionally, $ \alpha_y = \frac{CC_{y_r}}{CC_{y}} $, with the distances in y axis. The distances $w_i$ and $h_i$ are the width and height of the interface in pixels, respectively, and $w_m$ and $h_m$ are the markers size, width and height, respectively, in pixels. 

Finally, $p_t=(x_t, y_t)$ is the final point used to control the robot. The z coordinate and the robot's orientation are kept constant in this study.

\begin{figure*}[t]
    \centering
    \begin{minipage}{0.67\textwidth}
       \begin{minipage}{0.5\textwidth}
            \includegraphics[width=\linewidth]{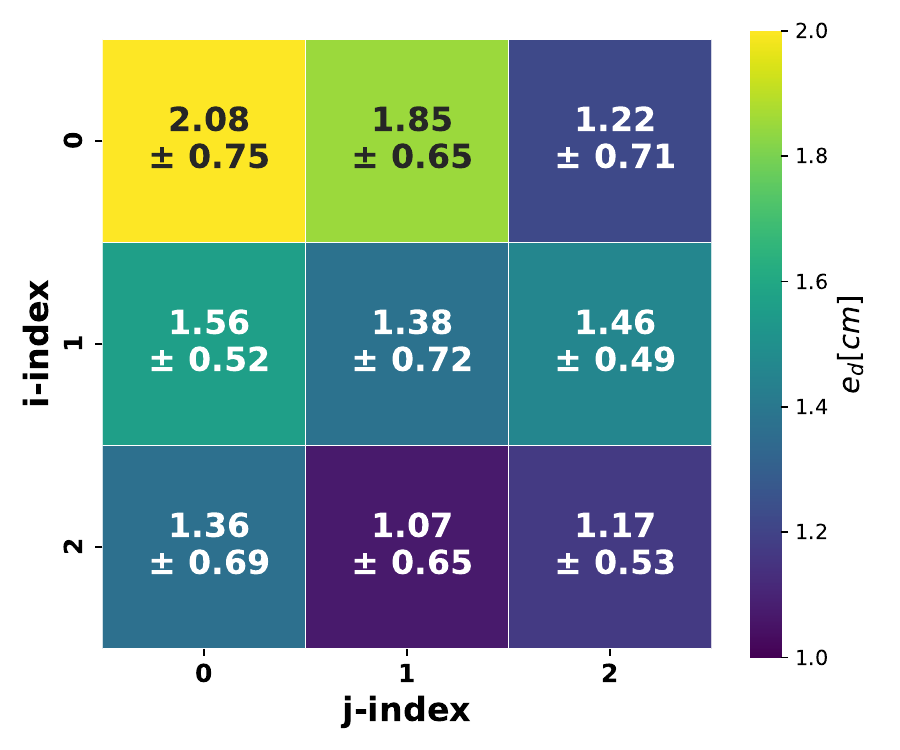}
            \text{a)}
            \label{fig:gaze-accuracy}
        \end{minipage}
        \begin{minipage}{0.5\textwidth}
            \includegraphics[width=\linewidth]{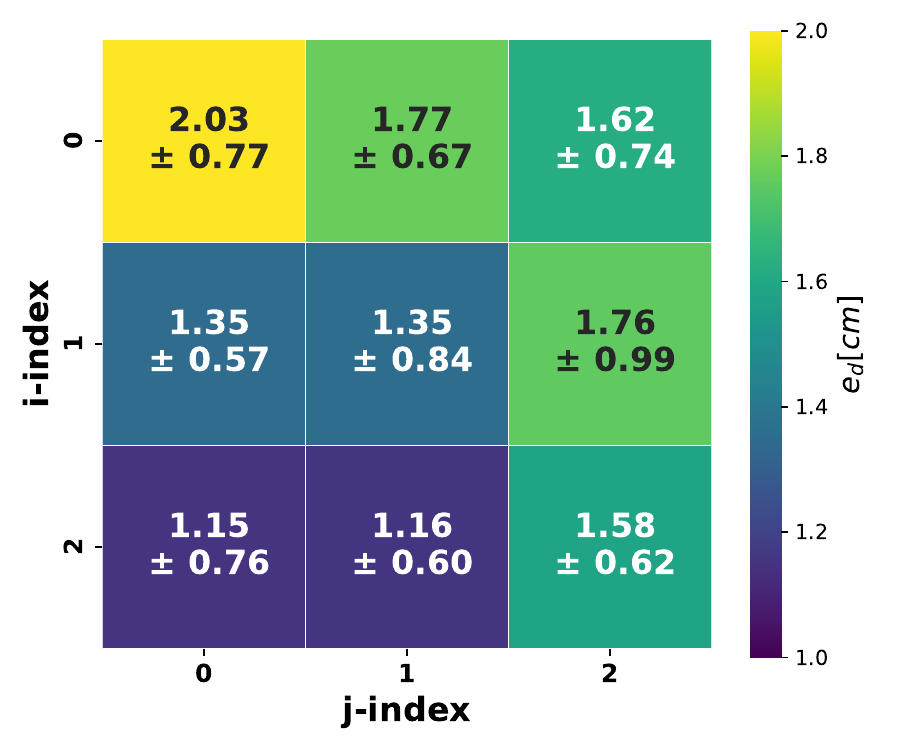}
            \text{b)}
            \label{fig:robot-accuracy}
        \end{minipage}
    \end{minipage}
    \begin{minipage}{0.32\textwidth}
        \begin{minipage}{\textwidth}
            \includegraphics[width=\linewidth]{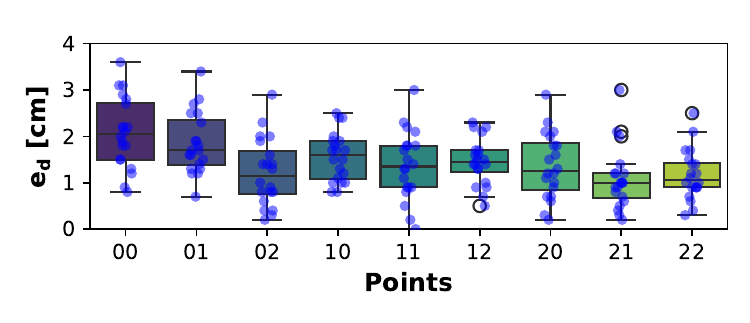}
            \text{c)}
            \label{fig:gaze-accuracy-box}
        \end{minipage}

        \begin{minipage}{\textwidth}
            \includegraphics[width=\linewidth]{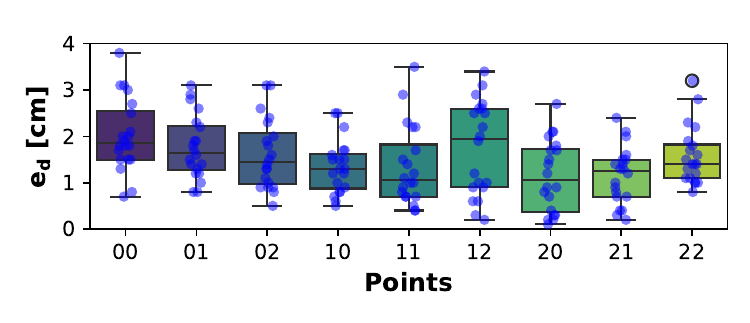}
            \text{d)}
            \label{fig:robot-accuracy-box}
        \end{minipage}
    \end{minipage}

    \caption{Test results for evaluating gaze fixation accuracy on the a) interface and b) robot. Each square in the heatmap represents a fixed point on the projected surface, displaying the average distance from the fixed point and its standard deviation. The results are also shown for data distribution as box plots for the c) interface and d) robot.} 
    \label{fig:accuracy}
\end{figure*}

\subsection{Evaluation of Gaze Accuracy on Interface}

An experimental setup with nine fixed points distributed across the surface where the interface is projected was developed to assess the accuracy of the user's fixed gaze.

Four participants\footnote{Authors of the paper. Three participants had visual impairments (myopia and astigmatism), while the other had no impairment.} were recruited to assess gaze accuracy. 
The subjects were asked to fix their gaze on the specified points as instructed in a randomized order, different for each subject, to avoid biasing. The protocol requires the user to first look at the center between the marker positioned at the left corner of the projection before each measurement and then direct their gaze toward the designated point, where this point was only disclosed to the user before the measurement. When the user fixes their gaze on the target for two seconds, the menu appears with a gray circle, with a white cross to mark its center, fixing the user's final point of interest. Each participant performed five measurements at each of the nine fixed points on the interface, totaling 45 individual measurements. This resulted in 180 measurements considering four subjects, 20 per point.

The error $e_d$ is measured directly on the surface in world coordinates between the user's intent and fixed point, which is given by the absolute Euclidean distance between the center of the fixed point $(x_{point}, y_{point})_{\Sigma}$ and the gray circle resulting from the user's intent $(x_{user}, y_{user})_{\Sigma}$, as shown in \eqref{eq:interface-error}. 

\begin{equation}
    e_d = \sqrt{(x_{point}-x_{user})^2+(y_{point}-y_{user})^2}
    \label{eq:interface-error}
\end{equation}

Before collecting the data, the user must calibrate the eye tracker by focusing on the center of a manufacturer's provided calibration card. To ensure consistency in environmental conditions, the user should position themselves in the location where the projection will take place, ensuring that the lighting remains constant throughout the experiment. The calibration target must be placed directly under the light emitted by the projector, ensuring proper calibration for accurate visualization in the given luminosity.

Results are shown in Fig.~\ref{fig:accuracy}a. The points closer to the user (e.g., on the third row) displayed the least mean error, as expected, since they had less distortion and noise, with only point 02 having a low mean error and being distant. The same analysis is applied to the most distant first row from the user, with the average gaze error. The overall accuracy is $1.46\pm0.71$ cm for the whole workspace, which is comparable to the results in the literature~\cite{Faisal2019, dziemian2016gaze}. 
In Fig.~\ref{fig:accuracy}c, the boxplot of the gaze errors for each point is displayed. The maximum error was measured on point 00, accounting for 3.65 cm, while the minimum was an exact gaze with 0.0 cm for point 20. 

It is also possible to see that besides 21 having the lowest mean and low standard deviation, it had a few outliers. This suggests that while the system generally performs well, there are occasional instances of significant deviation, which could be attributed to brief lapses in the user's focus or sudden head movements. Addressing these anomalies through adaptive algorithms that account for such variations could further enhance the system's reliability.

\subsection{Evaluation of Robot Position on the workspace}

After the previous experiment, the user was asked to select the option \textit{MOVE} in the menu to control the robot to the current gazed point.  
The robot arm, which had a pen attached to the end-effector, then executed its movement and descended near the projection surface. The absolute distance between the center of the fixed point on the projection and the robotic arm's final position was then measured.

The results are presented in Fig.~\ref{fig:accuracy}b. The robot had a similar error plot due to the input from the user already having an error during gaze. The overall error was $1.53\pm0.79$ cm. However, a different behavior can be seen in the third column, where the robot had a higher error. These errors are due to the imprecise positioning of the equipment, which led to cumulative errors in the experimental workbench's setup. Figure~\ref{fig:accuracy}d shows the error distribution, with a maximum of 3.85 cm at point 00 and 0.15 cm at point 20. 

Integrating feedback mechanisms that provide real-time corrections based on the robot's actual position relative to the target could enhance the robot's positioning accuracy. Integrating advanced sensors and machine learning algorithms to predict and compensate for potential errors in real time could also contribute to more precise and reliable robotic control.

\subsection{Evaluation of the Integrated Robotic System}

\begin{figure*}[t]
    \centering
    \begin{minipage}{\subfigwidth}
        \includegraphics[width=\linewidth]{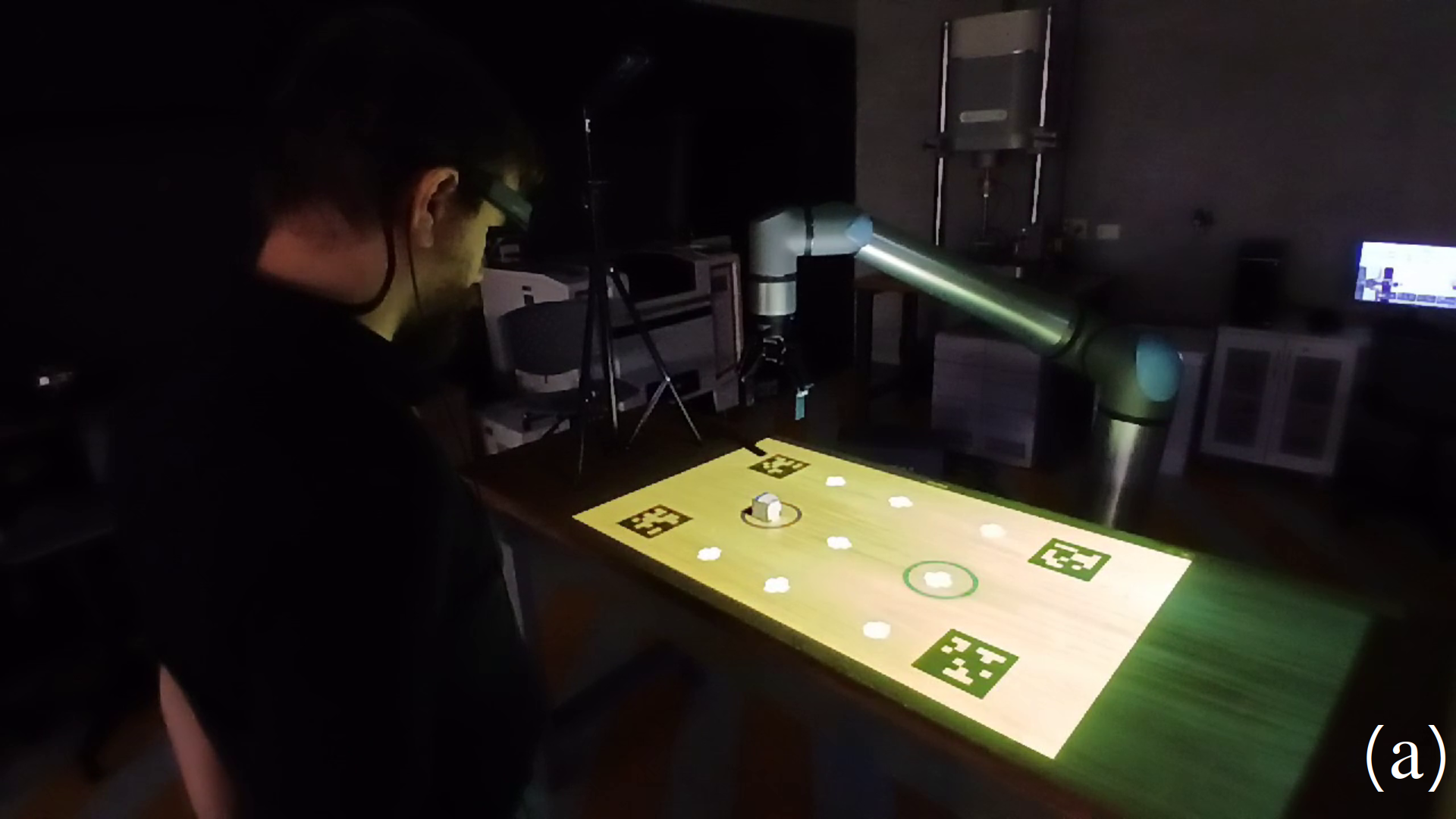}
    \end{minipage}
        \vspace{0.05cm}
    \begin{minipage}{\subfigwidth}
        \includegraphics[width=\linewidth]{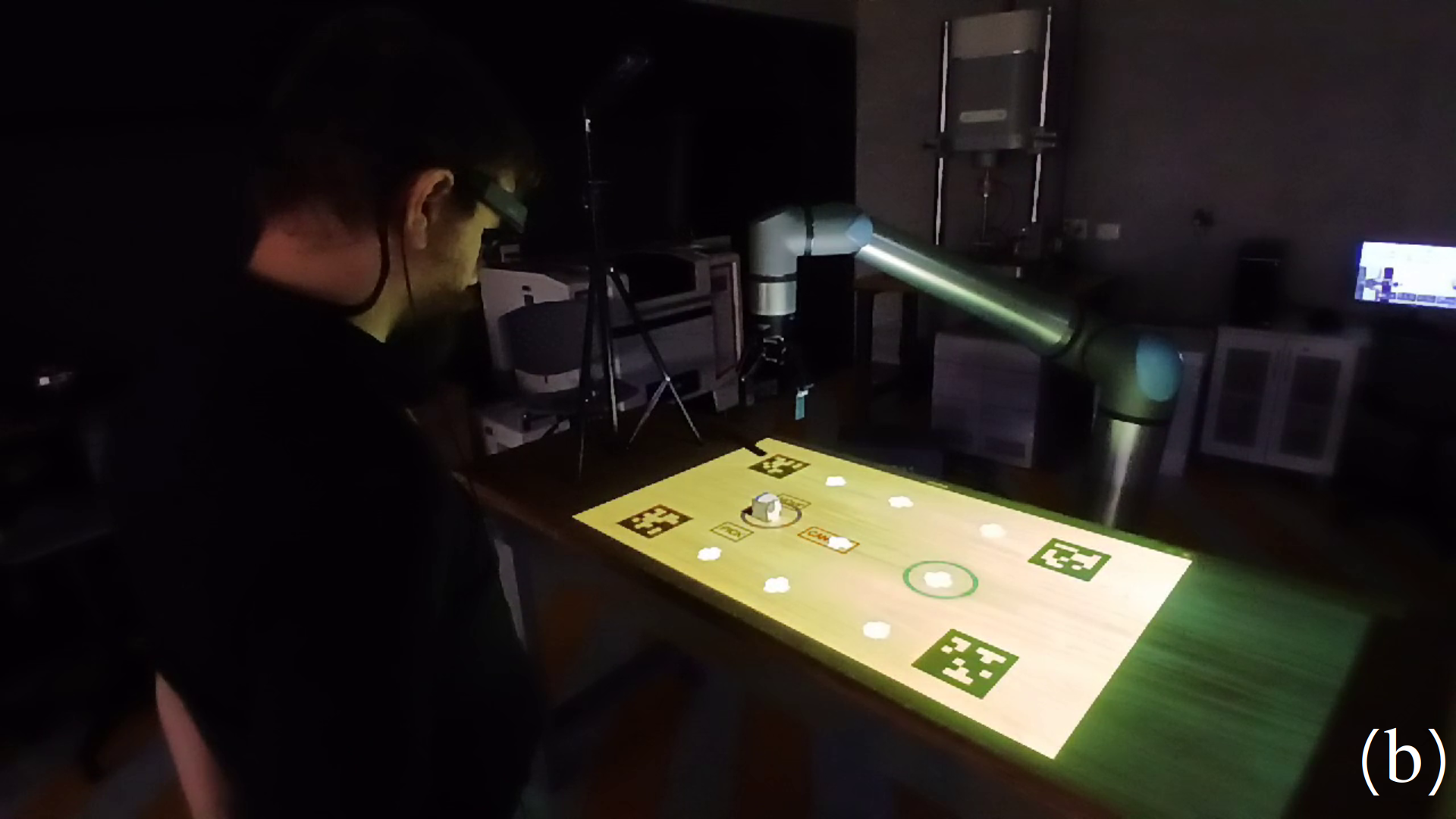}
    \end{minipage}
    \begin{minipage}{\subfigwidth}
        \includegraphics[width=\linewidth]{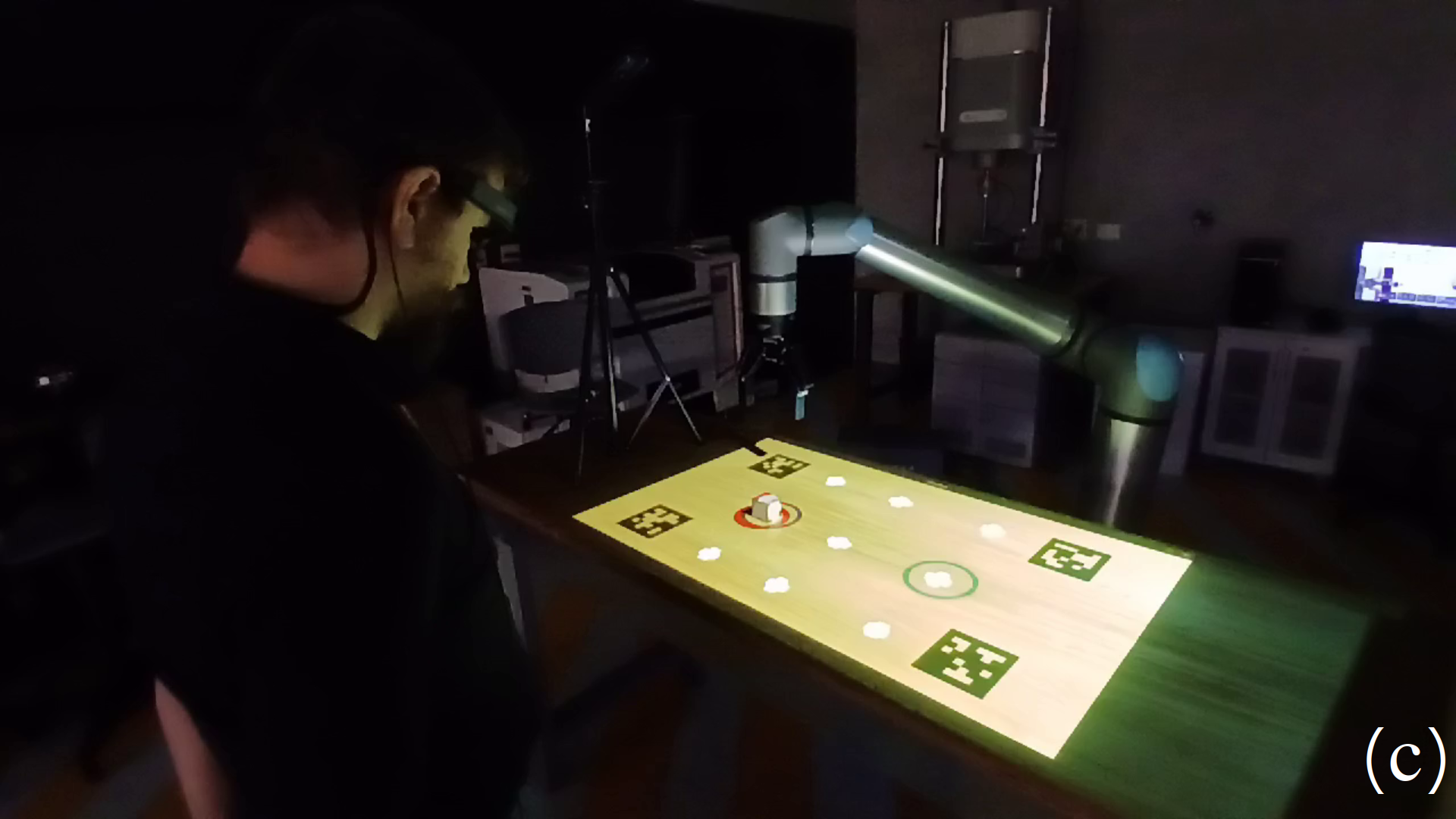}
    \end{minipage}
    \begin{minipage}{\subfigwidth}
        \includegraphics[width=\linewidth]{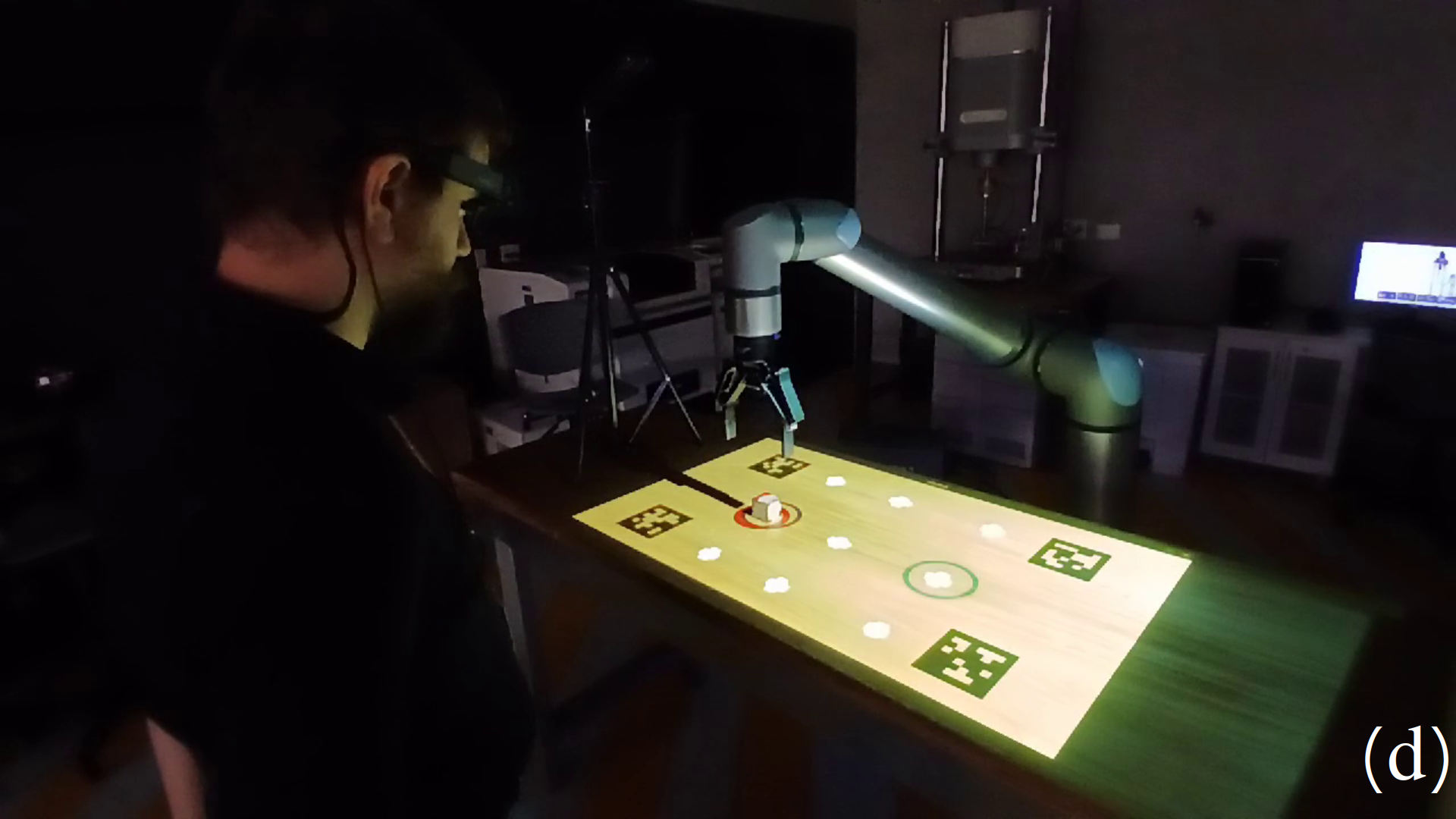}
    \end{minipage}
    
    \begin{minipage}{\subfigwidth}
        \includegraphics[width=\linewidth]{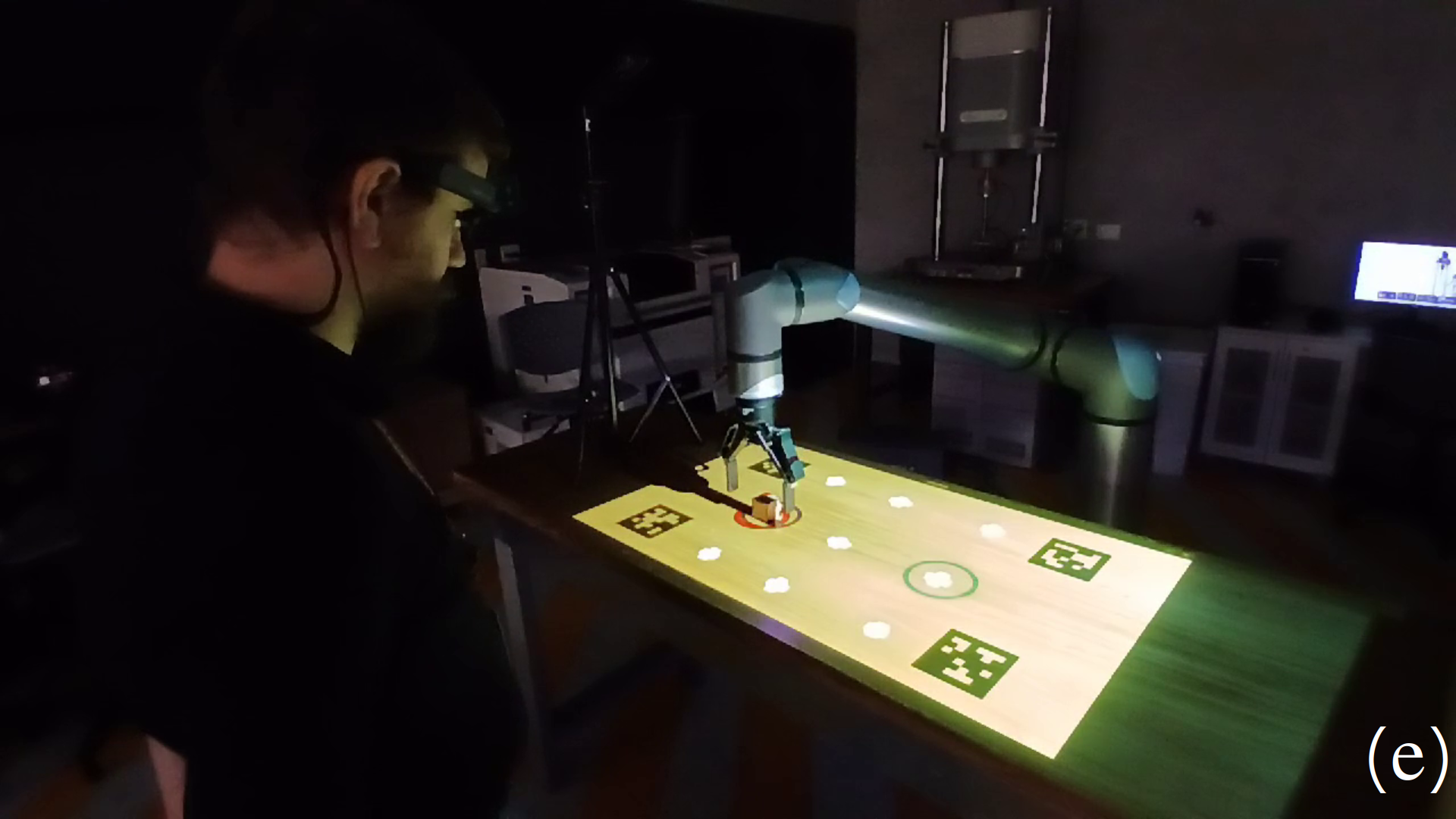}
    \end{minipage}
    \begin{minipage}{\subfigwidth}
        \includegraphics[width=\linewidth]{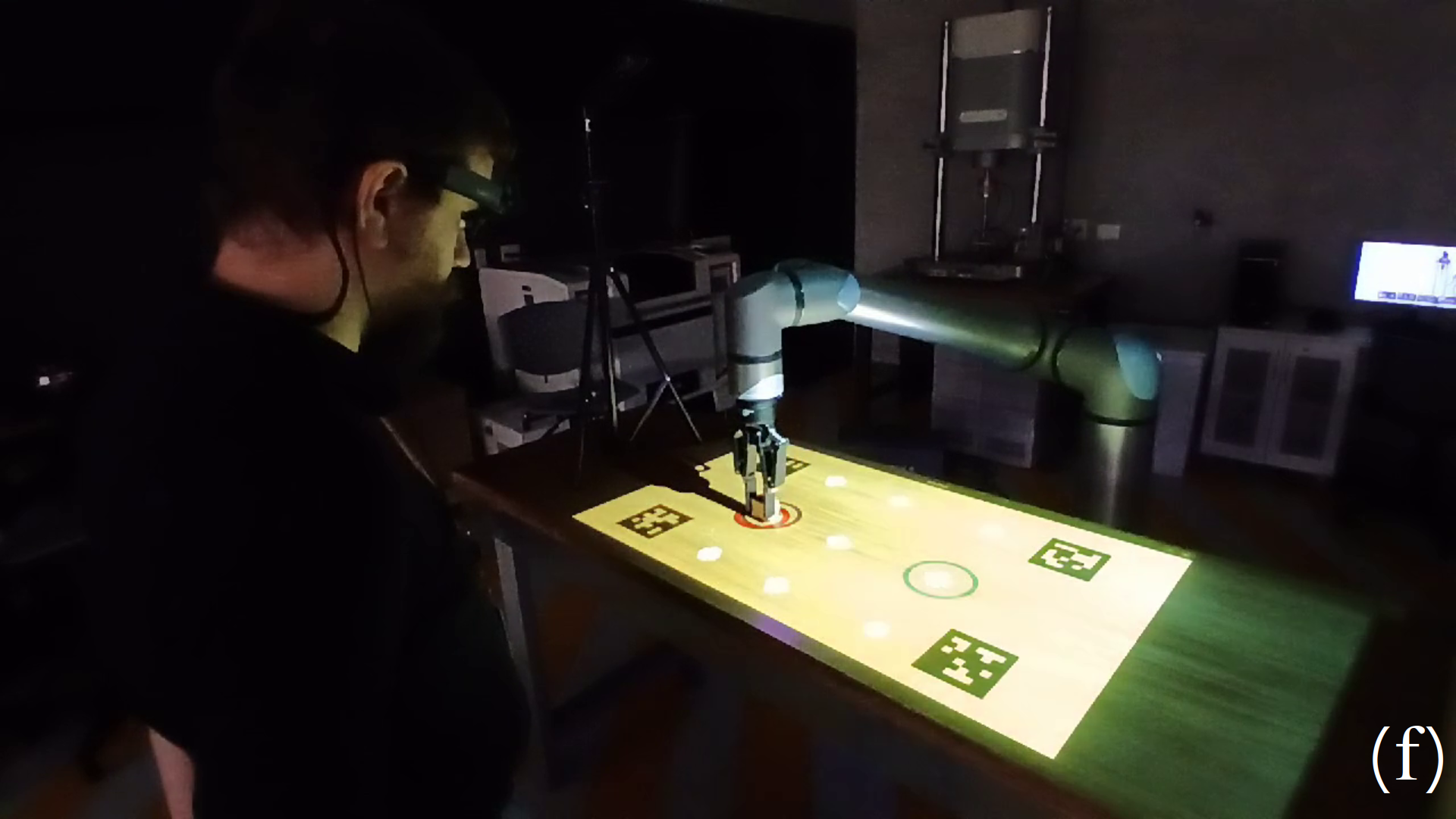}
    \end{minipage}
    \begin{minipage}{\subfigwidth}
        \includegraphics[width=\linewidth]{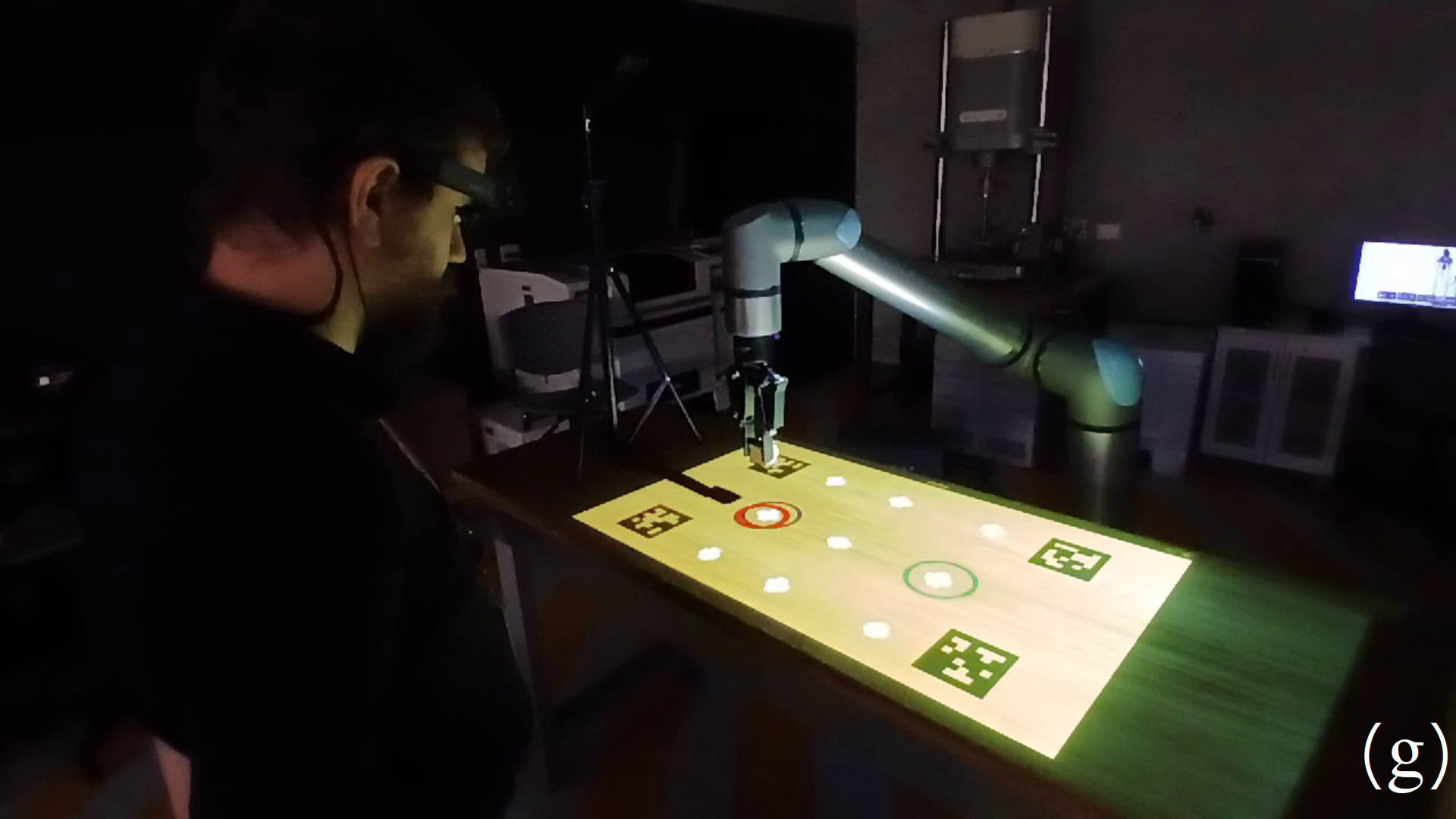}
    \end{minipage}
    \begin{minipage}{\subfigwidth}
        \includegraphics[width=\linewidth]{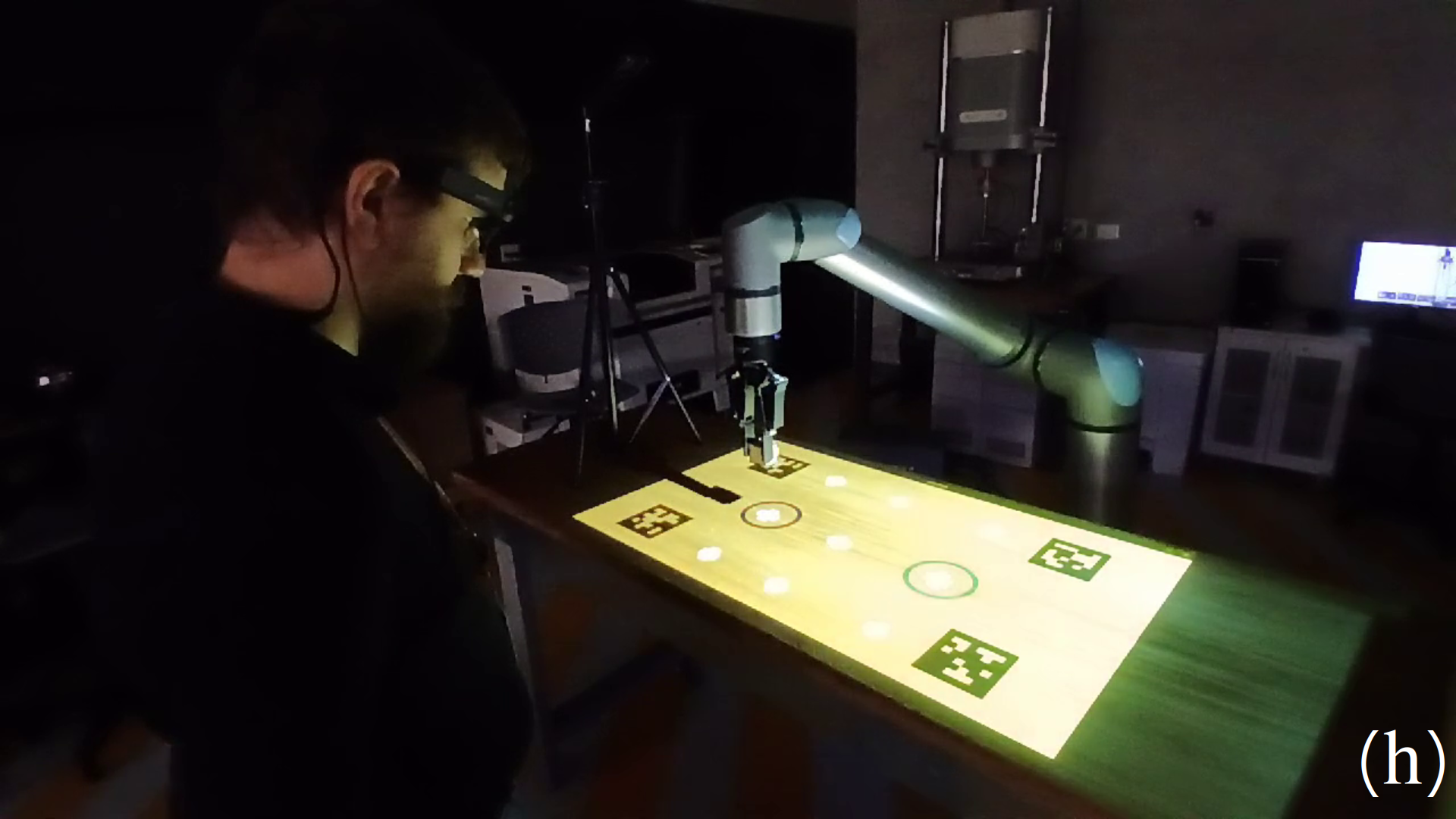}
    \end{minipage}
    
    \begin{minipage}{\subfigwidth}
        \vspace{0.05cm}
        \includegraphics[width=\linewidth]{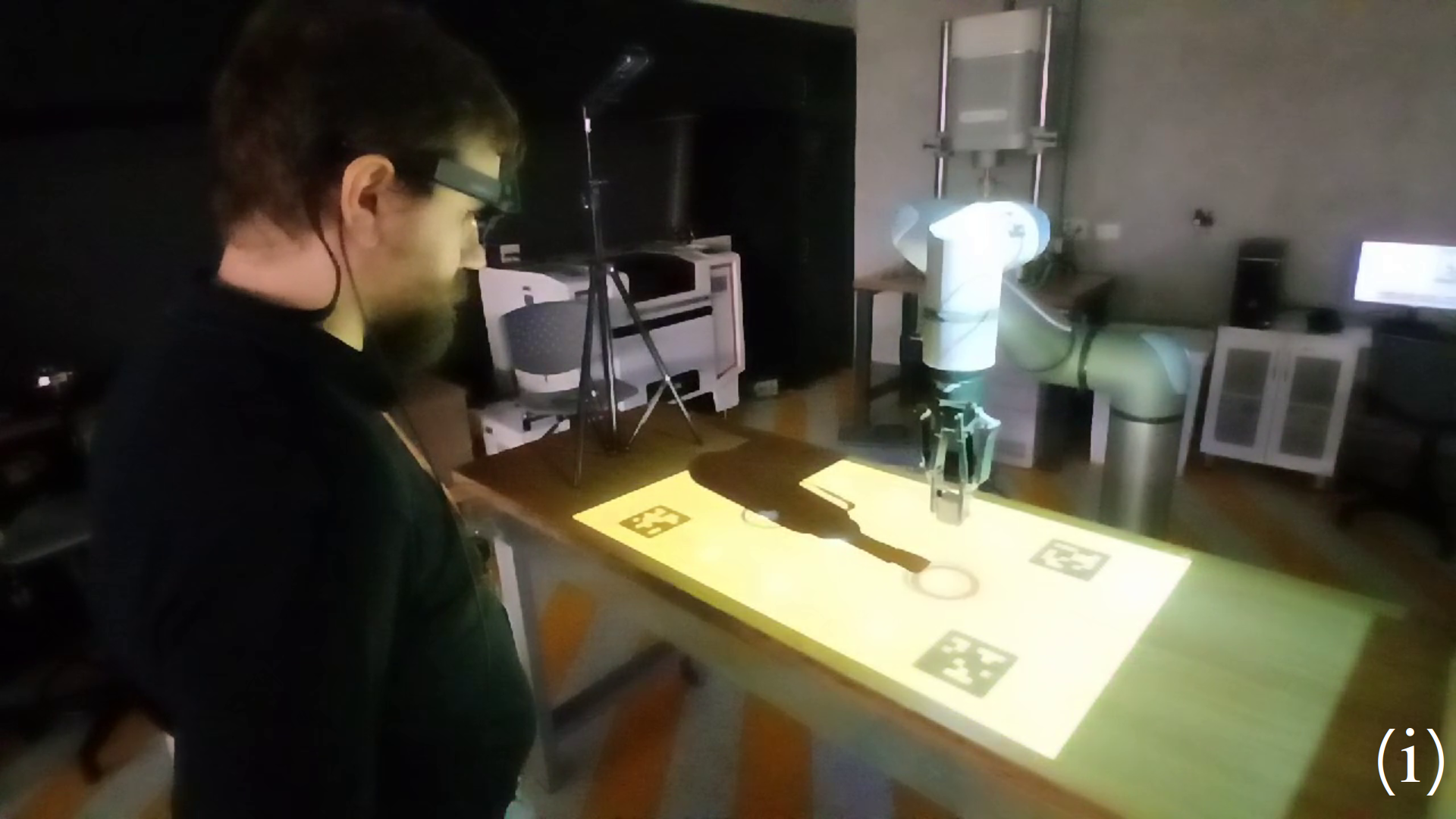}
    \end{minipage}
    \begin{minipage}{\subfigwidth}
        \includegraphics[width=\linewidth]{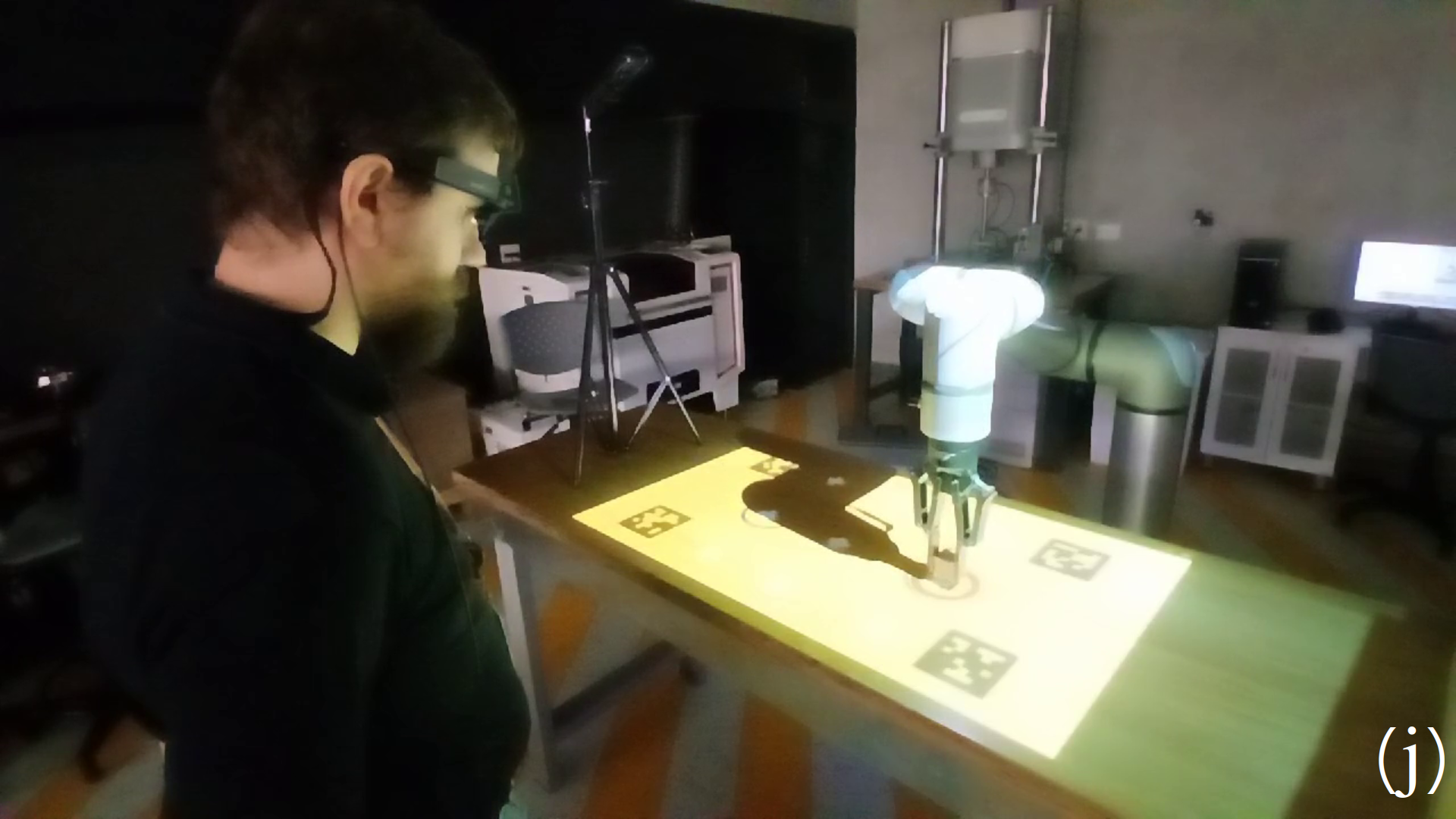}
    \end{minipage}
    \begin{minipage}{\subfigwidth}
        \includegraphics[width=\linewidth]{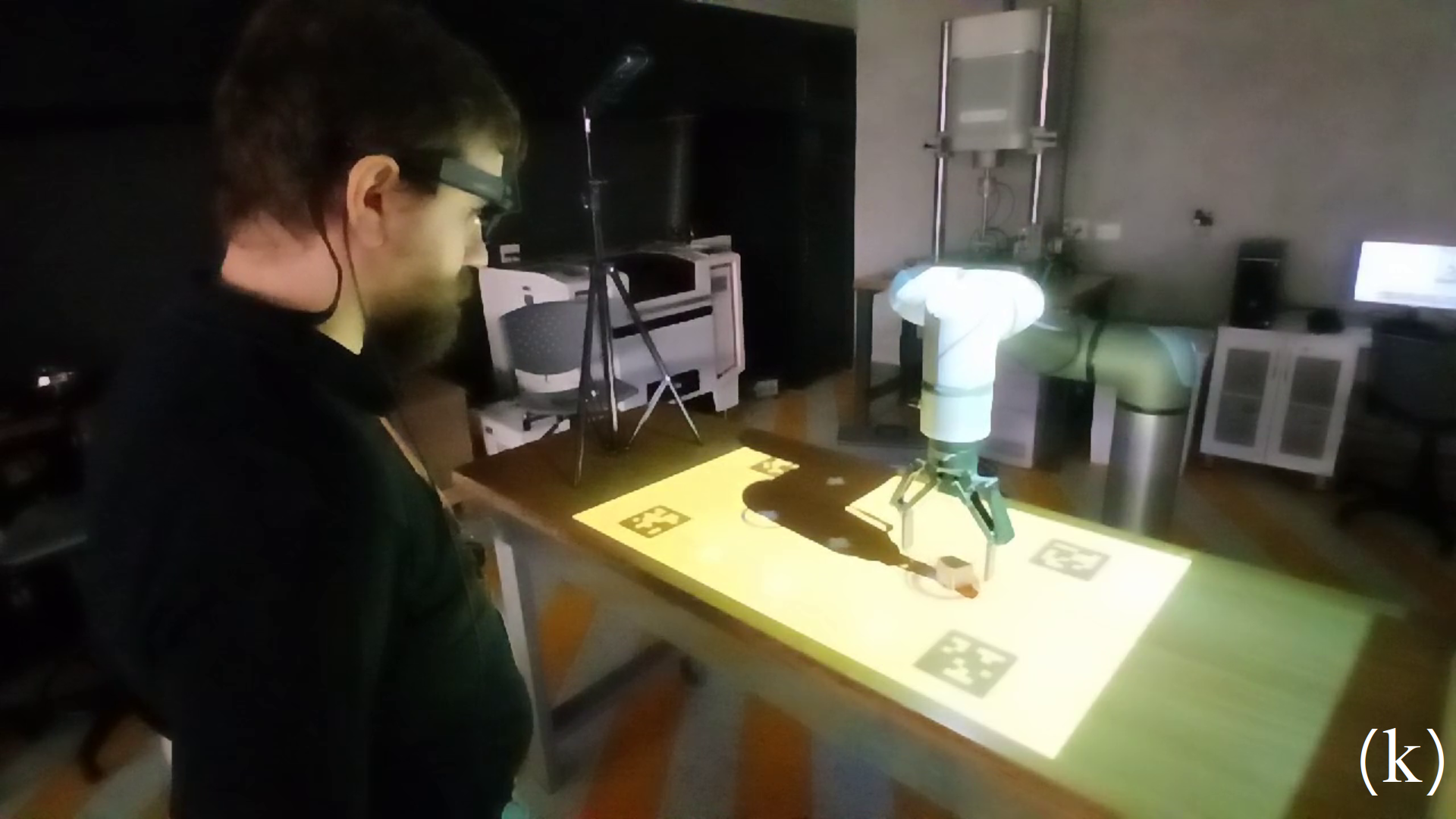}
    \end{minipage}
    \begin{minipage}{\subfigwidth}
        \includegraphics[width=\linewidth]{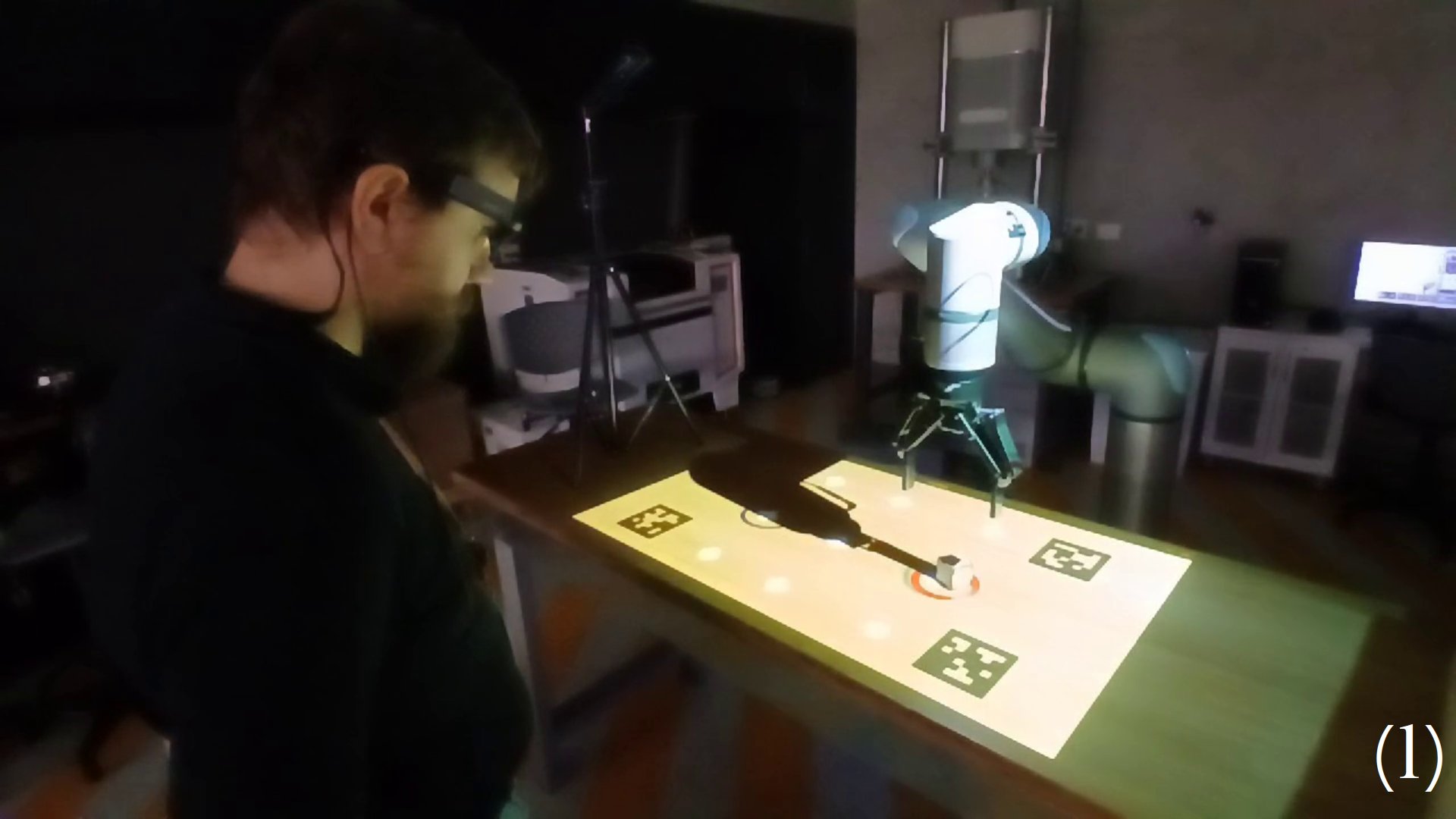}
    \end{minipage}

    \caption{Sequence of frames extracted from the video, illustrating key snapshots of the motion over time: From initial resting position (a), the user fixes his intention to the object position, opening the menu (b). Then, the user chooses the pick option, which the robot moves towards the object (c), stopping on the top (d), and moving down to grasp the object (e). Subsequently, the robot lifts the box (f) and waits for new commands (g). Now, the user tries to place the object in another interest area, fixing the gaze on the green circle (h). After selecting \textit{PLACE}, the robot moves to the new position (i), moves down to approach the interest area (j), places the box (k), and moves up to the waiting pose (l).}
    \label{fig:video_snapshots}
\end{figure*}

A pick-and-place experiment was proposed to validate the system's overall functionality. During the experiment, the user must perform the pick action by commanding the robot to grasp the object located at fixed point 10 (represented by the blue circle in Fig.~\ref{fig:intro}) and then execute the place action by positioning the block at fixed point 12 (represented by the green circle in Fig.~\ref{fig:intro}).

Measurements were collected from 12 points for each of the participating experimenters. The first two points were discarded to discard the learning phase. The success rate was defined based on the following conditions: the pick action was successful if the user command allowed the robot to grasp the object, while the place action was valid if the block was correctly positioned completely within the target point. The results of each repetition were recorded, with 1 representing success and 0 representing failure. If the pick action did not occur, the Place value was recorded as null. 

The overall success rate for picking the object achieved 95\% of success, while placing only 65\%. Several factors influenced the lower success rate during the place, despite solely inaccuracies from the gaze, which we name the main three. 

First, the robot did not put the object in contact with the workspace. Thus, it could bounce and end outside the circle when it released the object, marking it as failure. This could be improved by adding an interaction controller to guarantee contact with the table. Second, there were cumulative errors from the pick, where the object was successfully grasped but with the gripper closing closer to the corners instead of the edges. In this case, visual feedback would be needed, and it would be possible to correct the placement position. Instead of placing the end-effector center point in the desired position, the robot would use the object's geometrical center. Third, our gaze estimation currently does not consider the objects' depth. This could be improved by adding machine learning packages to deal with object recognition.

\section{CONCLUSIONS}

This paper proposes a novel solution combining gaze tracking, robotic assistance, and a mixed reality interface to enable mobility impaired individuals with intuitive hands-free object manipulation via robot control. By leveraging gaze-based selection mechanisms combined with real-time visual feedback, the system enhances user experience by improving clarity, reducing uncertainty, and increasing trust in the robot’s actions. Experimental evaluations demonstrated the effectiveness of this method, with a maximum average gaze error of 2.08 cm, an average robot positioning error below 2.03 cm, and a high task success rate in pick-and-place experiments.

The results highlight the potential of gaze-tracking and mixed-reality interfaces in assistive robotics, particularly for individuals with severe motor impairments. Integrating real-time feedback mechanisms improved system usability and reduced cognitive load, making it easier for users to adapt to gaze-based control. Furthermore, the findings suggest that removing real-time gaze feedback can reduce distractions and improve accuracy, emphasizing the need for carefully designed user interfaces in gaze-driven robotic systems.

Future work will focus on expanding the system’s capabilities, including adaptive calibration techniques for individual visual impairments, incorporating haptic feedback for enhanced user awareness, and integrating more complex object manipulation tasks. Additionally, conducting trials with a larger and more diverse user group will help further refine the system and validate its applicability in real-world assistive scenarios. This research represents the first step towards a promising field with several potentials, opening avenues for innovative advancements in human-robot interaction and assistive technologies.

\bibliographystyle{unsrt}  
\bibliography{references}

\end{document}